\pdfoutput=1
\def\year{2022}\relax
\documentclass[letterpaper]{article} 
\usepackage{amsmath}
\usepackage{aaai22}  
\usepackage{times}  
\usepackage{helvet}  
\usepackage{courier}  
\usepackage[hyphens]{url}  
\usepackage{graphicx} 
\usepackage{natbib}  
\usepackage{caption} 
\DeclareCaptionStyle{ruled}{labelfont=normalfont,labelsep=colon,strut=off} 
\usepackage{algorithm}
\usepackage{algorithmic}

\usepackage{booktabs}
\usepackage{multirow}
\usepackage{makecell}
\usepackage{amssymb}
\usepackage{siunitx}
\usepackage{subfig}

\usepackage{newfloat}
\usepackage{listings}
\floatstyle{ruled}
\newfloat{listing}{tb}{lst}{}
\floatname{listing}{Listing}
\def\vs{\emph{vs.}}
\def\eg{\emph{e.g.}}
\def\ie{\emph{i.e.}}

\title{OH-Former: Omni-Relational High-Order Transformer for Person Re-Identification}
\author{
    Xianing Chen\textsuperscript{\rm 1},
    Chunlin Chen\textsuperscript{\rm 1},
    Qiong Cao\textsuperscript{\rm 2},
    Jialang Xu\textsuperscript{\rm 3},
    Yujie Zhong\textsuperscript{\rm 4}, 
    Jiale Xu\textsuperscript{\rm 1},
    Zhengxin Li\textsuperscript{\rm 1},
    Jingya Wang\textsuperscript{\rm 1},
    Shenghua Gao\textsuperscript{\rm 1}
}
\affiliations{
    \textsuperscript{\rm 1}Shanghaitech University \\
    \textsuperscript{\rm 2}JD Explore Academy, JD.com\\
    \textsuperscript{\rm 3}University of Electronic Science and Technology of China\\
    \textsuperscript{\rm 4}Meituan Inc.\\
}

\usepackage{bibentry}

\begin{document}

\maketitle

\begin{abstract}

Transformers have shown preferable performance on many vision tasks. However, for the task of person re-identification (ReID), vanilla transformers leave the rich contexts on high-order feature relations under-exploited and deteriorate local feature details, which are insufficient due to the dramatic variations of pedestrians. In this work, we propose an Omni-Relational High-Order Transformer (OH-Former) to model omni-relational features for ReID. First, to strengthen the capacity of visual representation, instead of obtaining the attention matrix based on pairs of queries and isolated keys at each spatial location, we take a step further to model high-order statistics information for the non-local mechanism. We share the attention weights in the corresponding layer of each order with a prior mixing mechanism to reduce the computation cost. Then, a convolution-based local relation perception module is proposed to extract the local relations and 2D position information. The experimental results of our model are superior promising, which show state-of-the-art performance on Market-1501, DukeMTMC, MSMT17 and Occluded-Duke datasets.
\end{abstract}

\section{Introduction}
Person re-identification (ReID) aims to identify the same person from a set of pedestrian images captured by multiple cameras. This task is very challenging, since the attributes (\eg clothing, gender, hair) of pedestrians vary dramatically and their pictures are taken under various conditions (\eg illumination, occlusion, background clutter, and camera type). Hence, learning distinctive and robust features plays a decisive role in the field of person ReID.

CNN-based methods \cite{luo2019bag} have achieved great success in this field due to their strong ability in extracting deep discriminative features. In order to mine fine-grained local information from different body parts, part-based methods \cite{yao2019deep} are proposed to extract part-informed representations. By partitioning the backbone network's feature map horizontally into multiple parts \cite{fu2019horizontal, 2017Beyond} as shown in Figure 1(a), the deep neural networks can concentrate on learning more fine-grained salient features in each individual local part. 
However, due to the hard inductive bias of CNN, their models suffer from one common drawback, \ie, they require relatively well-aligned body parts for the same person. Besides, strict uniform partitioning of the feature map breaks with-part consistency. 

\begin{figure}[t]
	\centering
	\subfloat[Hand-craft Partition]{
		\centering
		\includegraphics[width=105pt,height=95pt]{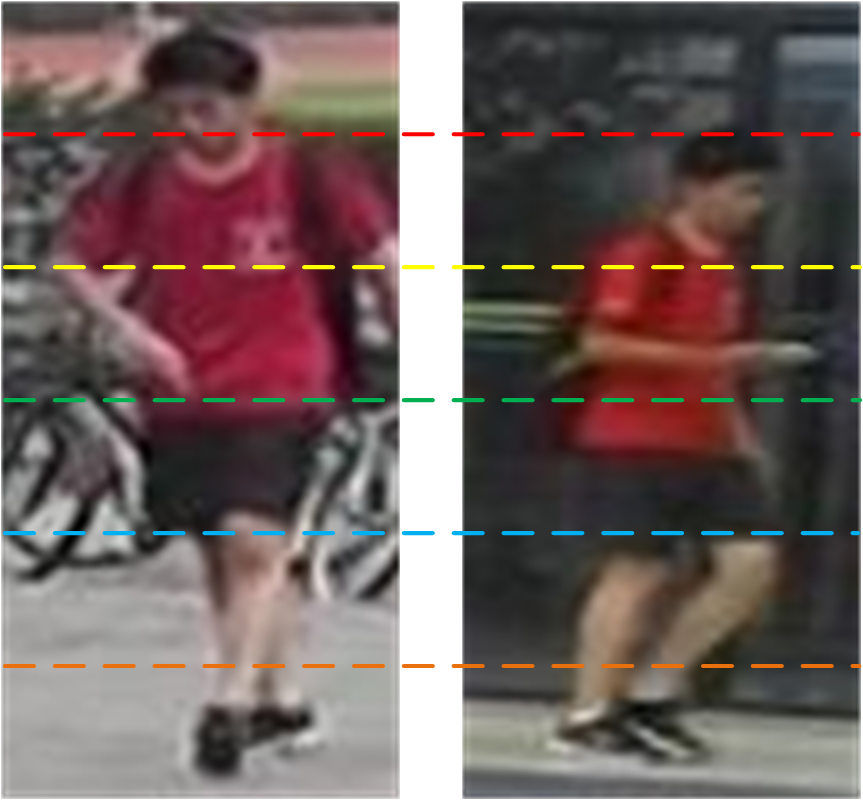}
		\label{fig1a}
	}
	\hfill
	\subfloat[Attention]{
		\centering
		\includegraphics[width=105pt,height=95pt]{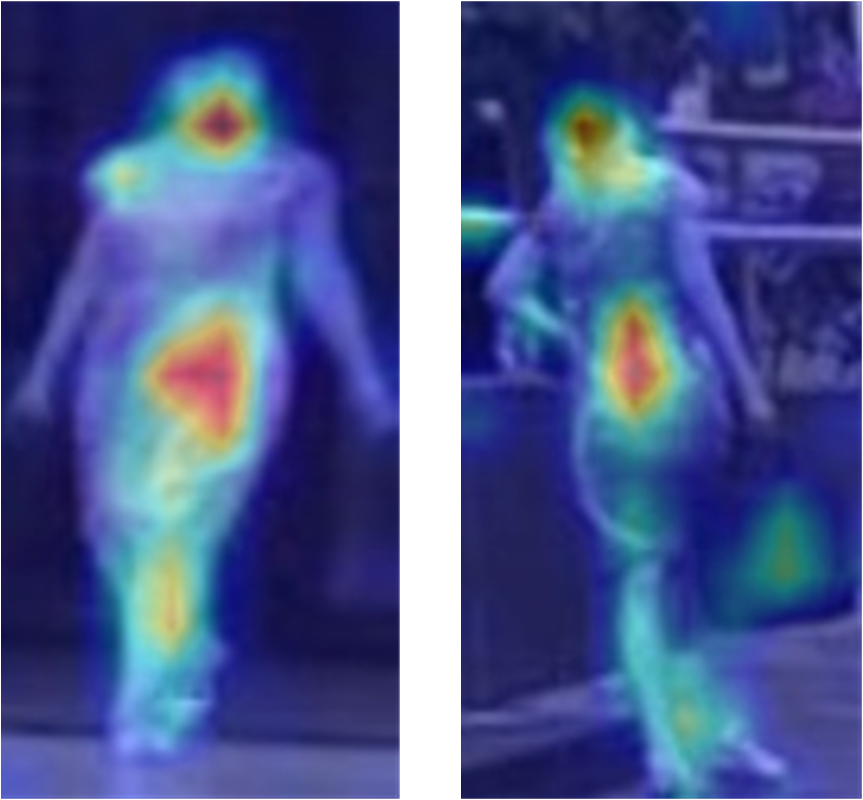}
		\label{fig1b}
	}
	\hfill
	\subfloat[Transformer]{
		\centering
		\includegraphics[width=105pt,height=95pt]{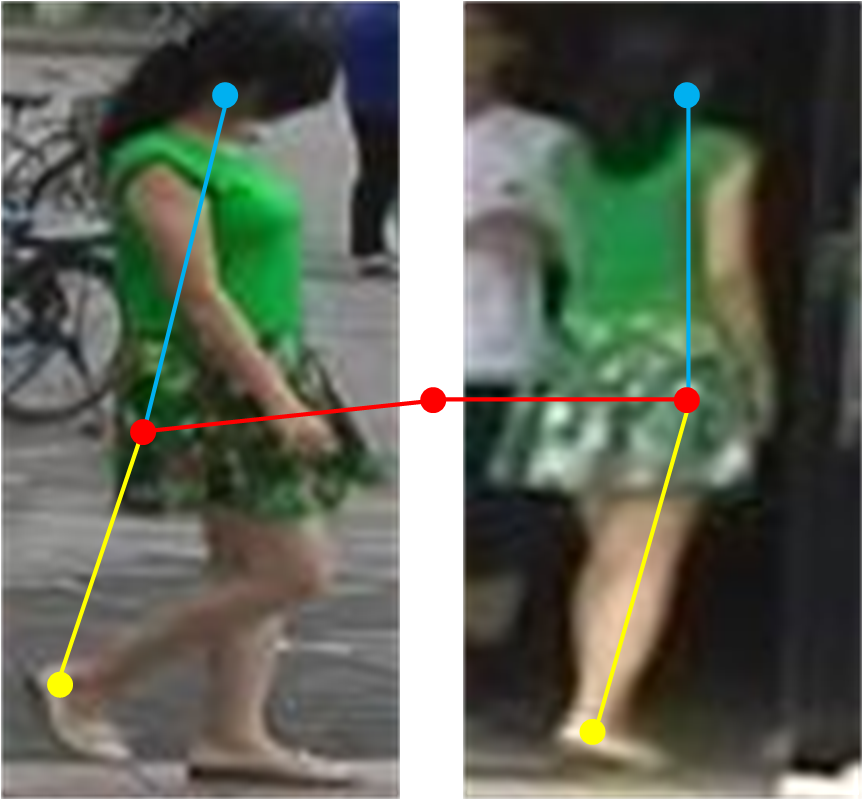}
		\label{fig1c}
	}
	\hfill
	\subfloat[Ours]{
		\centering
		\includegraphics[width=105pt,height=95pt]{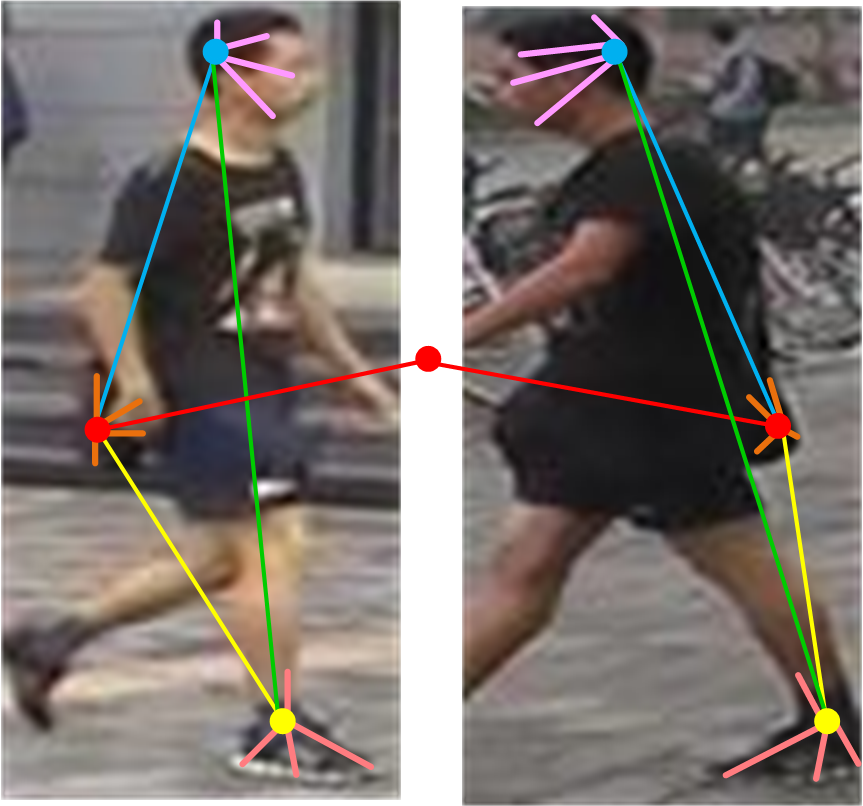}
		\label{fig1d}
	}
	\caption{Examples of different feature representation and matching methods for person ReID. (a) Hand-crafted partition methods require well-aligned body parts and breaks with-part consistency. (b) Attention-based methods consider only coarse region-level information. (c) Vanilla transformers ignore the rich contexts on high-order feature relations and local feature details. (d) Omni-relational high-order feature produced by our proposed OH-Former.}
	\label{fig1}
\end{figure}

Attention-based methods alleviate these problems \cite{cai2019multi} by locating human parts automatically as shown in Figure 1(b). However, they consider only coarse region-level attention whilst ignoring fine-grained pixel-level saliency and relational information.

Recently, with the success of Transformers in a variety of vision tasks, some previous work \cite{he2021transreid, li2021diverse} has attempted to introduce the power of self-attention mechanism into ReID.
By cutting the image into patches, transformers are capable of globally attending all the patches at every layer, making the spatial correspondences between input and intermediate features weaker. Nevertheless, most of the models obtain the attention matrices based on pairs of queries and isolated keys at each spatial location as shown in Figure 1(c), while leaving the rich contexts on high-order feature relations under-exploited. 
Moreover, although vision transformers can capture long-distance feature dependencies, they deteriorate local feature details \cite{peng2021conformer, li2021contextual}. This low-capacity representation is not robust to pedestrians with dramatic variations.
In addition, the performance of Transformers still lags behind CNNs in the low data regime \cite{dosovitskiy2020image} since vanilla Transformers lack certain desirable inductive biases processed by CNNs \cite{dai2021coatnet, d2021convit, guo2021cmt}, such as locality and weight sharing. These drawbacks clearly limit the application of Transformers on the ReID task. 

To solve these problems, we propose a novel Transformer-based model, \ie{Omni-Relational High-Order Transformer (OH-Former)}, which is equipped with a high-order transformer module with a prior mixing attention weight sharing mechanism and a local relation perception module (LRP). 
Specifically, we first obtain the first-order relation information by computing the non-local relations from pairs of queries and all other isolated keys at each spatial location, which named basic relations \cite{zhou2021graph}. 
Then we leverage the LRP which consists of a deformable and depth-wise convolution branch to dynamically model the local fine-grained feature relations and 2D spatial positions, which we call local relations. 
After that, by modeling the correlations of the positions in the feature map using non-local operations, the proposed module can integrate the local information captured by convolution operations and first-order long-range dependency captured by non-local operations. 
We call this robust feature representation as the omni-relational high-order feature, which contains both the high-order non-local information and the local detailed information as shown in Figure 1(d). For example, given a pedestrian image, there are some noticeable signals on blue and yellow dots for the red dot. After constructing the basic and local relations, the correlations among them are then captured by computing self-attention for the tokens. Using this mechanism, we explicitly tell the model that there are correlations among those tokens, the detailed information around the signals, the long-range fused information for those signals. Then the latter layers will learn under which circumstances such correlations are related to the identity information of the person.

High-order statistics can improve the model classification performance \cite{2018Towards, 2015Bilinear}, but they lead to high-dimensional representation and expensive computation cost simultaneously. Although the token number in our high-order layer is small enough after the aggregation of LRP, the $O(n^{2})$ non-local computation cost cannot be ignored. For efficient modeling, we propose a Prior Mixing Attention Sharing Mechanism for the high-order layer in OH-Former.
\citet{2019Sharing} found that good attention similarities exist among adjacent layers in Transformers. Intuitively, we want the high-order self-attention to extract high-order information for each location, thus the spatial correspondences can guaranteed for each order within a transformer layer. And we compute the Jensen-Shannon (JS) divergence to verify that each order within a layer actually generates similar attention weights. Therefore, we can just compute the weight matrix once in the first-order layer and reuse it in high-order layers. Moreover, pedestrians have some fixed and dominant patterns which can serve as a prior for identification. Different from hand-crafted attention prior \cite{2019Gaussian, 2018Modeling}, our shared attention is augmented by a learned prior to aggregate dominant and diverse information for high-order layer. 

In summary, the contributions of this paper are as follows:

(1) We propose a novel high-order transformer module that explores high-order relation information for person ReID. To the best of our knowledge, OH-Former is the first work to explore high-order information in Transformer.

(2) We propose a Prior Mixing Attention Weights Sharing Mechanism to reduce the computation cost and model the dominant and diverse features.

(3) To incorporate with the inductive biases of CNNs, we design a local relation perception module (LRP) to extract and aggregate the local information.

(4) Comprehensive experiments on four person ReID benchmark datasets demonstrate that our proposed model OH-Former achieves state-of-the-art performance.

\section{Related Works}
\subsection{Person Re-identification}
Several person ReID methods based on CNNs have recently been proposed. To extract fine-grained features from different body parts, they often utilize part-based methods \cite{ yao2019deep} to enhance the discriminative capabilities of various body parts. The fine-grained parts are usually generated by roughly horizontal stripes \cite{2017Beyond, fu2019horizontal, wang2018learning, he2018deep}. They require relatively well-aligned body parts for the same person. However, person pictures localized by the off-the-shelf object detectors often have the spatial semantics misaligned problem. Other methods make use of external cues such as pose \cite{zheng2019pose, miao2019pose, wang2020high, gao2020pose}, parsing \cite{kalayeh2018human, he2019foreground, he2020guided} information to align body region across images. These approaches usually require extra sub-networks and computation cost in inference and the accuracy is limited to the power of the estimator.  Attention-based methods \cite{li2018harmonious, liu2017hydraplus, cai2019multi} learn feature representations robust to background variations and focus on most informative body parts, however, they are not capable of modeling fine-grained information and relations. Thus, transformer-based methods are introduced to explore the power of self-attention to model robust person feature representation.

\subsection{Vision Transformers}
With the success of Transformers \cite{vaswani2017attention} in natural language processing, many studies \cite{dosovitskiy2020image,  chen2021pre, touvron2021training} have shown that they can be applied in computer vision as well. Transformer-based methods have boosted various vision tasks such as image classification \cite{yuan2021tokens}, object detection \cite{carion2020end, zhu2020deformable}, and segmentation \cite{xie2021segformer}. Recently, some work \cite{he2021transreid, li2021diverse, wu2021person} explores the power of transformers in the field of ReID. However, they do not consider the subtle high-order representation differences among pedestrians which is missing in the vanilla transformers.

More importantly, transformers show preferable performance on large datasets while lagging behind CNNs in the low data regime since they lack the inductive biases which is inherent in CNNs \cite{dosovitskiy2020image, dai2021coatnet}, \eg locality and translation equivariance. To solve this problem, previous work \cite{peng2021conformer, li2021contextual, guo2021cmt, d2021convit, li2021localvit} attempts to introduce convolutions to Transformers to employ all the benefits of CNNs while keeping the advantages of Transformers. However, their designs only introduce 2D locality but ignore local feature similarity, \ie, local relations. It will damage Transformers' representation obtained from self-attention. In contrast, our model is aware of local relations by augmenting the spatial sampling locations with learned offsets \cite{dai2017deformable}.

\subsection{High-Order Information}
The high-order statistics have been successfully exploited in fined-grained visual classification tasks \cite{zhou2021graph, 2018Towards, 2015Bilinear, 2017Is}.
For the ReID task, \citet{chen2019mixed} models high-order attention statistics information to capture the subtle differences. \citet{2021Coarse} introduces a second-order information bottleneck to cope with redundancy and irrelevance features.  \citet{Bryan2019Second} models local features' long-range relationships. \citet{2020High} learns high-order topology information for occluded ReID. However, non of them consider the high-order statistics of non-local information which is important for modeling pedestrians. We instead model an omni-relational feature to capture both the local and non-local relation differences among pedestrians.

\begin{figure}[t]
	\centering
	\includegraphics[width=0.99\columnwidth]{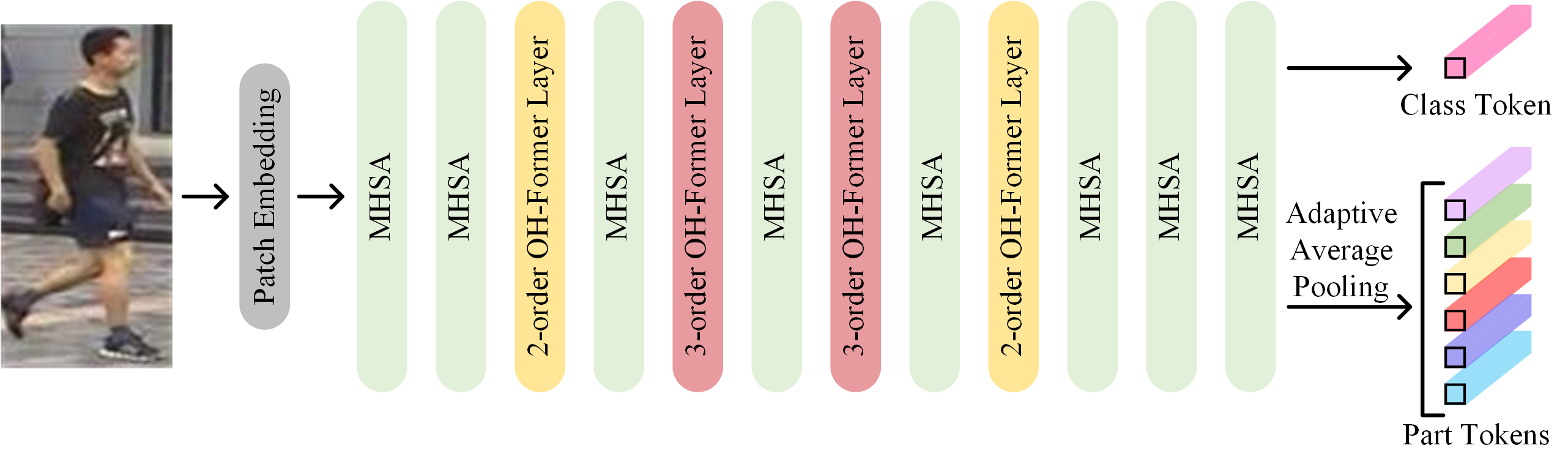}
	\caption{The overall architecture of our OH-Former for ReID. We embed an image into a sequence of flattened tokens, concatenate a class token, add position embeddings and feed them into stacked MHSA and our OH-Former layers to exploit the omni-relational high-order information. The output tokens without the class token are fed into an adaptive average pooling layer to get part tokens. Finally, we use the class and part tokens to train classifiers or do inference.}
	\label{fig2}
\end{figure}

\section{Proposed Method}
\subsection{Overall Architecture}
The overview of OH-Former is presented in Figure 2. Given an image of $x \in R^{H \times W \times C}$, where $H, W, C$ denote the image's height, width, and channel number, respectively. We utilize the stem architecture as our patch embedding layer which has a $5 \times 5$ convolution with stride  5 to reduce the feature size, followed by a $3 \times 3$ convolution with stride 2 for better local information extraction \cite{guo2021cmt} to get a 2D feature $x \in R^{h \times w \times c}$. Then we flatten the feature into a sequence of tokens $X \in R^{n \times d}$, where $n=h \times w$ is the resulting number of patches and $d$ is the channel number. Following the design of \cite{dosovitskiy2020image}, we concatenate a class token to those tokens and add a learnable position embedding on them. 
Afterwards, we process those tokens $X \in R^{T \times d}$ with stacked Multi-head Self-Attention (MHSA) and our OH-Former layers. The resulting tokens without class token $z_{cls}$ are passed to an adaptive average pooling layer \cite{2017Beyond} to get part tokens $z_{i} (i=1,2, ..., p)$, where $p$ is the number of parts. 

Finally, we optimize the network by minimizing the sum of Cross-Entropy and Triplet losses with BNNeck \cite{luo2019bag} over the class and part tokens. Specifically, our model is optimized with the following losses:
\begin{equation}
\begin{split}
L = L_{CE}(z_{cls})+L_{Triplet}(z_{cls}) + \\
      \frac{1}{p}\sum_{i=1}^{p}(L_{CE}(z_{i})+L_{Triplet}(z_{i})) 
\end{split}
\end{equation}
where triplet loss is computed as $L_{Triplet}=[d_{p}-d_{n}+\alpha]$. $d_{p}$, $d_{n}$ are feature distances of positive pair and negative pair, respectively. $\alpha$ is the margin parameter.

During inference, we concatenate the class and part tokens to form the final feature representation and compute the Euclidean distances between them to determine the identities of different people.

\begin{figure}[t]
	\centering
	\includegraphics[width=0.95\columnwidth]{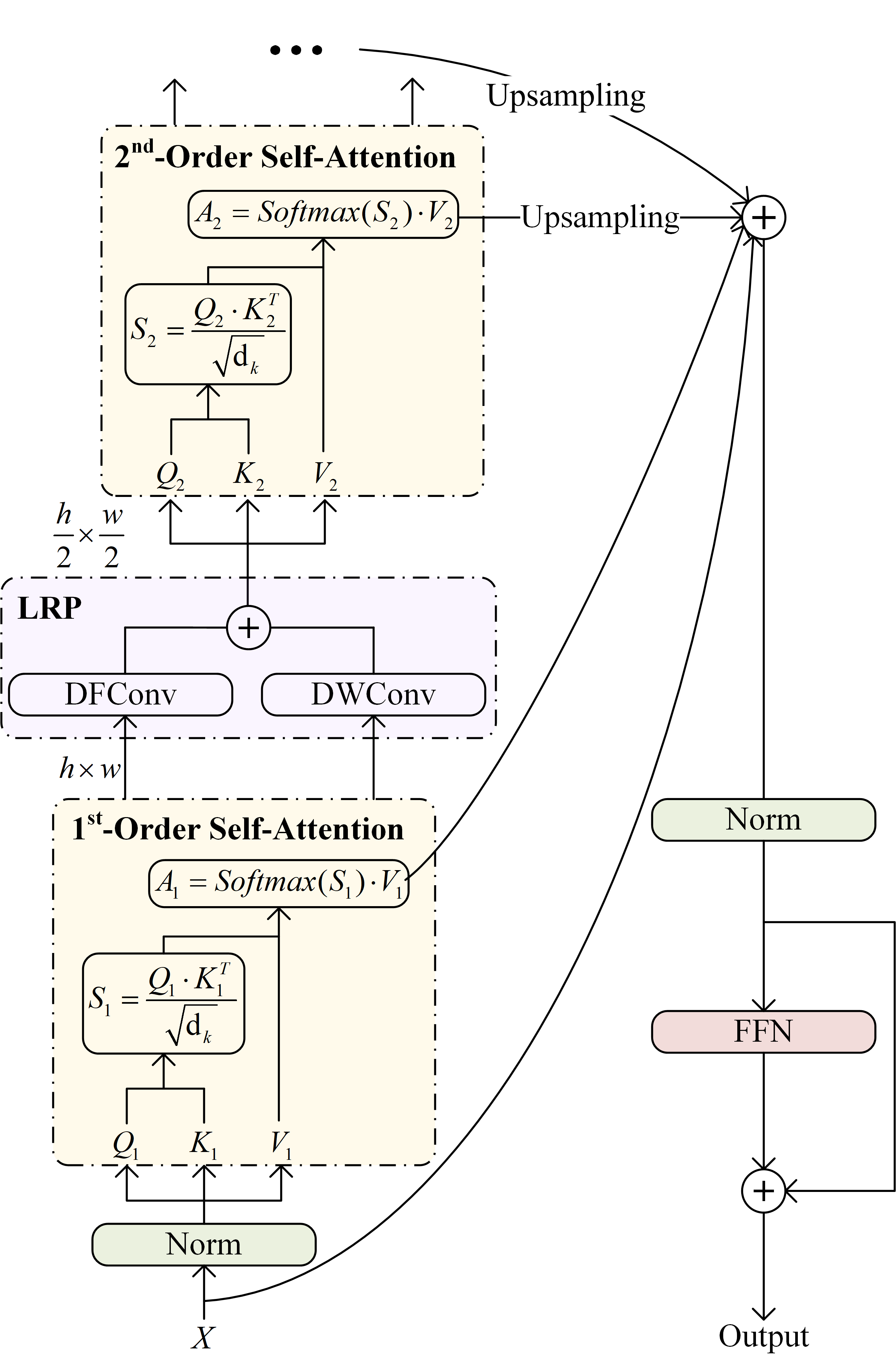}
	\caption{Detailed structure of our OH-Former layer.}
	\label{fig3}
\end{figure}

\subsection{Omni-Relational High-Order Transformer}
In this section, we will describe our proposed OH-Former layer in details.
\subsubsection{First-Order Self-Attention.}
Given an input sequence $X \in R^{T \times d}$, the first-order self-attention performs a scaled dot product attention following the vanilla Transformer \cite{vaswani2017attention, dosovitskiy2020image}, defined as:
\begin{equation}
S_{1}=\frac{Q_{1}\left(K_{1}\right)^{T}}{\sqrt{d_{k}}}, 
s.t. Q_{1}=X W^{Q_{1}}, K_{1}=X W^{K_{1}}
\end{equation}
\noindent where  $W^{Q_{1}} \in R^{d \times d_{k}}, W^{K_{1}} \in R^{d \times d_{k}}$ are linear transformations. $S$ is an $l \times l$ matrix, entry $(i, j)$ represents the similarity between position i and position j. For simplicity, the multi-head self-attention mechanism is omitted here.

The output of the first-order self-attention is defined as the weighted sum of values:
\begin{equation}
A_{1}=Softmax\left( S_{1} \right) \times V_{1}, s.t. V_{1}=X W^{V_{1}}
\end{equation}
\noindent where $V_{1}$ is generated from the same source with a linear transformation $W^{V_{1}} \in R^{d \times d_{v}}$. There basic relations \cite{zhou2021graph} construct the one-\vs-rest relations \cite{park2020relation} of body tokens.

\subsubsection{Local Relation Perception Module.}
Although vision transformers can capture long-distance feature dependencies, they ignore the local feature details \cite{yuan2021tokens}. Fortunately, convolutions's hard inductive biases encoded by locality and weight sharing can compensate for this drawback \cite{peng2021conformer}. Previous methods have introduced inductive bias to Transformers by fixed grid depthwise convolution \cite{li2021contextual, li2021localvit, dai2021coatnet}, which are unaware of local relations. They actually bring in 2D positions to transformers \cite{xie2021segformer}. 
In contrast, we propose a Local Relation Perception Module (LRP), which consists of deformable \cite{dai2017deformable} and depthwise convolution \cite{tan2019efficientnet} branches as shown in Figure 3. 
The deformable convolution branch dynamically aggregates local relation information across patches by attending to sparse spatial locations, producing the refined local-relation-aware feature. The depthwise convolution branch uses a fixed $3 \times 3$ depthwise convolution with stride 2 to extract positional information \cite{chu2021conditional, xie2021segformer} for high-order features. Specifically, LRP is defined as:
\begin{equation}
LRP(A_{i}) = DeformConv(A_{i}) + DWConv(A_{i})
\end{equation}

\noindent where $A_{i}$ is the $i$-th order feature. $DeformConv(\cdot)$ denotes the deformable convolution and $DWConv(\cdot)$ denotes the depth-wise convolution. Without loss of generality for other Transformer variants, we feed the first-order features without class token to our LRP. For simplicity, the reshape operations are omitted here. 

\subsubsection{Omni-Relational High-Order Self-Attention.} After capturing the basic and local relations, we take a step further to model our omni-relational high-order features.
For constructing the relations among basic and local relations, we use multi-head self-attention to process features processed by LRP. Given the i-th order features $A_{i}$, the (i+1)-th order information $A_{i+1}$ is extract by 
\begin{equation}
S_{i+1}=\frac{Q_{i+1}\left(K_{i+1}\right)^{T}}{\sqrt{d_{k}}} 
\end{equation}
\begin{equation}
A_{i+1}=Softmax\left(S_{i+1}\right) \cdot V_{i+1} 
\end{equation}
\noindent where $Q_{i+1}=A_{i} W^{Q}, K_{i+1}=A_{i} W^{K}, 	V_{i+1}=A_{i} W^{K}$. After that, $A_{i+1}$ is processed by LRP again for computing higher-order self-attention. 

Compared to OH-Former, vanilla transformers can also capture high-order statistics, but they need to stack many layers which leads to high computation cost, and deep transformers are very hard to train.   

\begin{figure}[t]
	\centering
	\includegraphics[width=0.95\columnwidth]{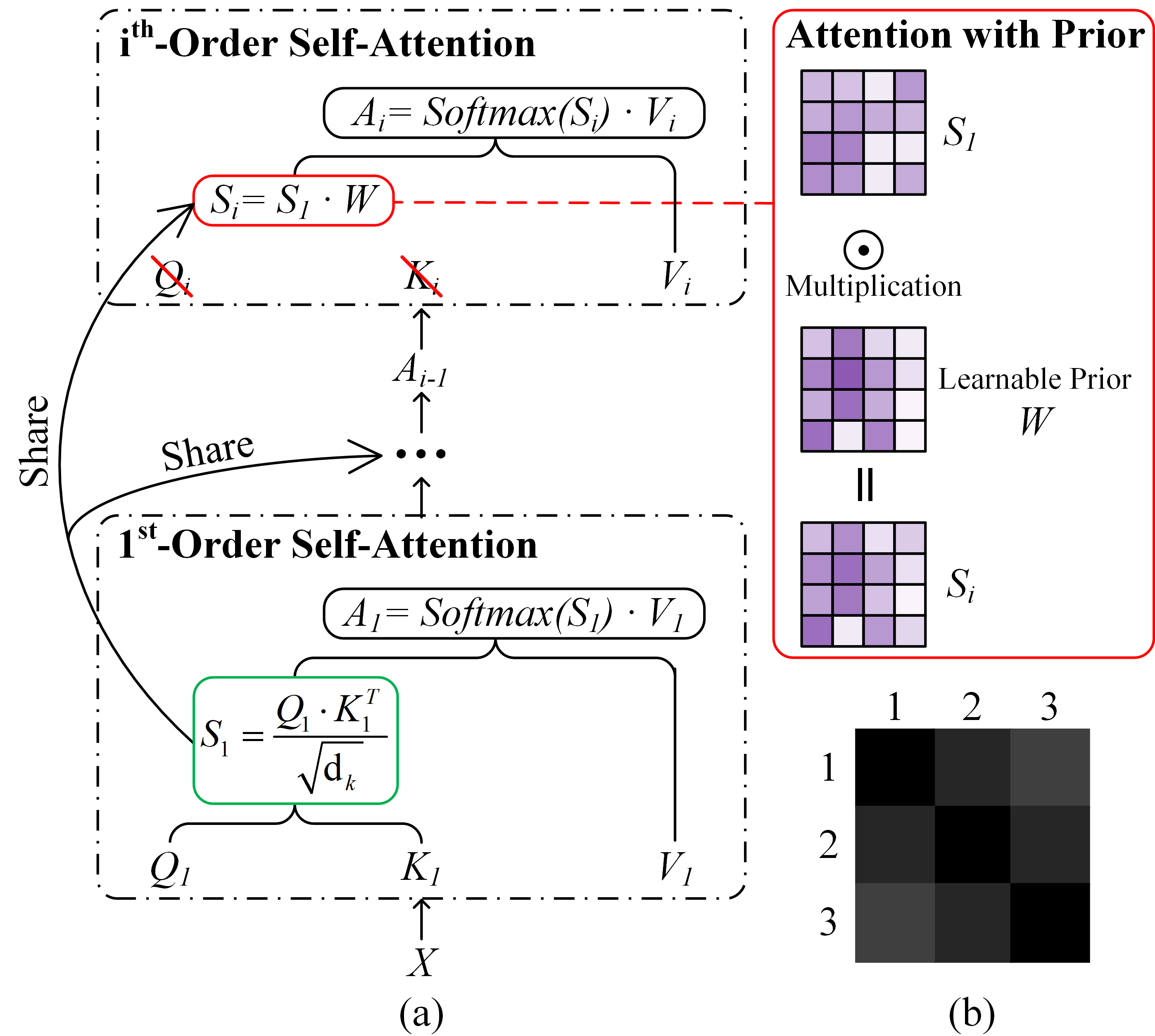}
	\caption{Illustration of our proposed mechanism. (a) Process of our Prior Mixing Attention Sharing Mechanism. (b) Jensen-Shannon divergence of the attention weights for different orders within a OH-Former layer (Darker cell indicate a more similar distribution). We use a three-order OH-Former layer as an example.}
	\label{fig4}
\end{figure}

\subsubsection{Prior Mixing Attention Sharing Mechanism.}
Although the number of tokens in high-order layer is small enough after being aggregated by LRP, the $O(n^{2})$ self-attention computation cost is still too high. For efficient modeling, we propose a Prior Mixing Attention Sharing Mechanism for each order within an OH-Former layer. The idea is that we only compute the weight matrix once and reuse it in high-order layers. 

Intuitively, we want the high-order self-attention to extract high-order information for each location, so the spatial correspondence relationship should be guaranteed for each order within an OH-Former layer. And we compute JS divergence to measure the similarity between the high-order and the first-order self-attention weights. Figure 4(b) shows that the model generates similar weights over orders. Thus, we can share $S_{1}$ without class token similarity to high-order layers:
\begin{equation}
S_{i}=S_{1}, i=2, ...
\end{equation}

Moreover, person images have some fixed pattern for identification which can serve as a prior. To aggregate dominant and diverse information, we propose a prior mixing mechanism. After computing attention from inputs (\eg $Softmax(QK^{T})$), the attention is further augmented by a learned prior. Specifically, the Prior Mixing Attention Sharing Mechanism is defined as:
\begin{equation}
S_{i}=S_{1} \cdot W, i=2, ...
\end{equation}
\noindent where $W$ is the learnable parameters.

\subsubsection{Omni-Relational Feature Fusion Feed-Forward Network.} The relations modeled by each order layer are already omni-relational high-order relations. We then fuse them in an effective way. For each high-order feature $x \in R^{T \times d}$, we reshape the feature to $x \in R^{h \times w \times d}$ and use the nearest interpolation to upsample it to $x \in R^{H \times W \times d}$ which has the same shape as the reshaped first-order feature without class token. Then we flatten it,  concatenate it with a zero-value vector $z_{0} \in R^{1 \times d}$ and sum it with the first-order feature. Finally, the fused feature is fed into a Layer-Normalization layer followed by a Feed-Forward layer with skip connection.

\subsubsection{Model Variations.} Although the vanilla transformers can learn high-order statistics in the high-level layer by stacking non-local operations and they need to extract simple context patterns in the low-level layer \cite{d2021convit}, the order setting of OH-Former is  still flexible. So we show our insight on this in the ablation studies chapter.

Additionally, we build two variants of our model, named OH-Former and OH-Former$_{Share}$. OH-Former$_{Share}$ uses our prior mixing attention sharing mechanism to share and re-calibrate attention weights of different orders to trade-off model performance and computational complexity. 

\begin{table*}[!ht]
	\centering
	\setlength{\tabcolsep}{17pt}
	\begin{tabular}{ccccccc}
		\toprule[1pt]
		\multirow{2}{*}{Method} 	&\multicolumn{2}{c}{DukeMTMC-reID}
		&\multicolumn{2}{c}{Market1501}
		&\multicolumn{2}{c}{MSMT17} \\
		\cmidrule(lr){2-3}	\cmidrule(lr){4-5}	\cmidrule(lr){6-7}	
		&mAP	&R-1	&mAP	&R-1	&mAP	&R-1  \\
		\midrule
		HACNN \cite{li2018harmonious}&63.8 &80.5 &75.7 &91.2 &- &-\\
		PCB \cite{2017Beyond}&66.1 &81.7 &77.4 &92.3 &- &-\\
		CBN \cite{2020Rethinking}&67.3	&82.5 &77.3	&91.3 &42.9 &72.8\\
		PCB+RPP \cite{2017Beyond}&69.2 &83.3 &81.6 &93.8 &-	&-\\ 
		HPM \cite{fu2019horizontal}&74.3 &86.6 &82.7 &94.2 &- &-\\ 
		GASM \cite{he2020guided}&74.4 &88.3 &84.7 &95.3 &52.5 &79.5\\ 
		CTF \cite{2021Coarse}&74.9 &87.4 &87.7 &94.8 &- &-\\
		SAN \cite{2020Semantics}&75.7 &87.9 &88.0 &96.1   &55.7 &79.2\\
		BoT \cite{luo2019bag}&76.4 &86.4 &85.9 &94.5 &-   &-\\
		MHN \cite{wang2018learning}&77.2 &89.1 &85.0 &95.1 &- &-\\
		MGN \cite{wang2018learning}&78.4 &88.7 &86.9 &\textbf{95.7} &52.1 &76.9\\
		SCSN \cite{2020Salience} &79.0	&\textbf{91.0} &88.5	&\textbf{95.7}	&58.5	&83.8\\  
		ViT-BoT \cite{he2021transreid}	&79.3 &88.8 &86.8 &94.7 &61.0   &81.8	\\
		ISP \cite{2020Identity}&80.0 &89.6	&88.6	&95.3	&-	&-\\
		TransReID \cite{he2021transreid}&82.6 &90.7  &\textbf{89.5} &95.2 &\textbf{69.4} &86.2\\
		OH-Former$_{Share}$ (Ours) &81.9 &90.2	&88.0	&94.9	&67.6 & 85.6 \\
		OH-Former (Ours) &\textbf{82.8} &\textbf{91.0} &\underline{88.7}	&95.0 &\underline{69.2}	&\textbf{86.6}\\
		\bottomrule[1pt]
	\end{tabular}
	\caption{Performance comparision with state-of-the-art methods in holistic ReID. The best two results are in bold and underline.}
	\label{tab4}
\end{table*}

\section{Experiments}
\subsection{Datasets}
We evaluate our model on the following publicly available ReID datasets and compare its performance with state-of-the-art methods. 1) The Market1501 dataset \cite{2016Scalable} containing \num{32668} person images of \num{1501} identities. 2) The DukeMTMC-reID \cite{2016Performance} composed of \num{36441} images of \num{1404} identities. 3) The MSMT17 \cite{2018Person} containing \num{126441} images of \num{4101} identities. 4) The Occluded-Duke dataset \cite{miao2019pose} with occluded person images selected from DukeMTMC-reID.

\begin{table}[!htbp]
	\centering
	\setlength{\tabcolsep}{5pt}
	\begin{tabular}{lcc}
		\toprule[1pt]
		LRP Type	&mAP	&R-1	\\
		\midrule
		None &80.1	&89.4 \\
		DWC	 &81.7	&90.2 \\
		NC	 &81.8	&90.1 \\
		AP	 &80.7	&89.9 \\
		MP	 &80.4	&89.7 \\
		DWC + AP &81.8 &90.2 \\
		AP + DFC &82.3  &90.8 \\
		DWC + DFC (Ours) &\textbf{82.8}	&\textbf{91.0}\\
		\bottomrule[1pt]
	\end{tabular}
	\caption{Performance comparison with different variants of LRP on DukeMTMC-reID dataset. DWC: Depthwise Convolution; AP: Average Pooling; MP: Max Pooling; NC: Normal Convolution; DFC: Deformable Convolution. None means we do not add any downsampling operation after the self-attention operation.}
	\label{tab1}
\end{table}

\subsection{Implementation}
\subsubsection{Training.} All person images are resized to $368 \times 128$. The training images are augmented with random horizontal flipping and random erasing. The batch size is set to \num{256} with \num{16} images per ID. Stochastic gradient descent optimizer (SGD) is employed with momentum of \num{0.9} and weight decay of \num{1e-4}. The learning rate is initialized as \num{0.01} with cosine learning rate decay. For the MHSA and first-order self-attention layers, we use the imagenet-pretrained model of ViT-B \cite{dosovitskiy2020image}. Our model is implemented with PyTorch library~\cite{PyTorch} and trained on four Nvidia Titan X GPUs.

\subsubsection{Evaluation Protocols.} We use standard metrics in ReID community, namely Cumulative Matching Characteristic (CMC) curves and mean Average Precision (mAP), to evaluate all methods. We report ReID results under the setting of a single query for a fair comparison.

\subsection{Ablation Study of LRP}
The effectiveness of the proposed LRP Module is validated in Table \ref{tab1}. We compare the performance of several variants of our LRP in terms of mAP and rank-1 accuracy. From the first three rows, we can see that convolution improves the retrieval performance by introducing 2D position information. Although its fixed grid convolution kernel can capture local information, it is unaware of local relation. Thus, for dynamically aggregating high-order information which has complex relations aggregated by non-local operations, a local-relation-aware operation is critical to our model. Also, the worst result in the first row demonstrates that without local downsampling operations the high-order self-attention will learn homogeneous information and the effective of our LRP is significant. From the fourth, fifth and seventh rows, we show that compared with convolution, pooling operations will lead to the loss of large amount of information, This is due to the fact that local information in non-local operations is more complex than it is in CNN features, not only because pooling is an parameter-free operation. The second, sixth and last rows show exploring local relations to construct our omni-relational feature is important, and the performance reaches to the peak when using all components.



\begin{table}[htbp]
	\centering
	\setlength{\tabcolsep}{2.5pt}
	\begin{tabular}{ccccccc}
		\toprule[1pt]
		OH-Former order &mAP &R-1\\
		\midrule
		$[None]$ &78.6  &88.9 \\
		$[H_2^{0,1}]$ &78.5	&88.9 \\
		$[H_2^{9,10,11}]$ &78.8 &89.0	\\
		$[H_2^{2,4,6,8}]$ &82.4	&90.9 \\ 
		$[H_3^{2,4,6,8}]$ &\textbf{83.0} &\underline{91.2} \\
		$[H_4^{2,4,6,8}]$ &\underline{82.9}	&\textbf{91.3} \\
		$[H_2^{2,8},H_3^{4,6}]$	&82.8 &91.0 \\
		\bottomrule[1pt]
	\end{tabular}
	\caption{Ablation results for the OH-Former  with different orders on DukeMTMC-reID dataset. $H_i^j$ means we use an i-order OH-Former layer at the $j$-th layer. Note that we only show OH-Former layers for simplicity.}
	\label{tab2}
\end{table}

\subsection{Ablation Study of OH-Former}
We demonstrate the effectiveness of the proposed OH-Former layer in Table \ref{tab2}. 
From the second row, we can observe that high-order information will slightly decrease the performance since Transformers need to learn simple context patterns on the low level. The third row shows that high-order statistics can only brings a small performance boost since vanilla Transformers can learn high-order statistics in the high level layers by stacking transformer layers. Thus, the model benefits more in the middle levels as shown in the last five rows. And the order higher than three in OH-Former layer just delivers a negligible boost as shown in the sixth row. As a result, we choose the configuration $[H_2^{2,8},H_3^{4,6}]$ in the last row to trade off speed and performance.

Ablation experiments from overall and individual perspectives are also conducted on our proposed method. Our OH-Former layers bring 5.34\% mAP improvement over the baseline by modeling omni-relational high-order features. OH-Former achieves the best performance, which shows that advantages of LRP and OH-Former layer are complementary and their coherent innovations contribute to a high-performance ReID model. The forth row shows that Attention Weights Sharing will decrease the model performance. But after adding our Attention Prior (PM), the model shows comparable performance to our OH-Former model by fusing dominant and diverse information. 

\begin{table}[htbp]
	\centering
	\setlength{\tabcolsep}{1.5pt}
	\begin{tabular}{lcccccc}
		\toprule[1pt]
		Method	&LRP	& OH & AS & PM	& mAP	&R-1	\\
		\midrule
		Baseline	&$\times$	&$\times$	&$\times$ &$\times$	&78.6  &88.9 \\
		OH-Former w/o LRP &$\times$	&\checkmark	&$\times$ &$\times$ &80.1 &89.4 \\
		OH-Former &\checkmark	&\checkmark	&$\times$	&$\times$	&\textbf{82.8} &\textbf{91.0} \\
		OH-Former$_{Share}$ w/o PM	&\checkmark	&\checkmark	&\checkmark	&$\times$	&80.3 &90.0 \\
		OH-Former$_{Share}$	&\checkmark	&\checkmark	&\checkmark	&\checkmark	&\underline{81.9} &\underline{90.2} \\
		\bottomrule[1pt]
	\end{tabular}
	\caption{Ablation results for the proposed OH-Former on DukeMTMC-reID dataset. LRP: Local Relation Perception module for OH-Former layer. OH: OH-Former layer. AS: Attention Sharing. PM: Prior Mixing.}
	\label{tab3}
\end{table}

\subsection{Comparison with State-of-the-Art Methods}
\subsubsection{Results on Holistic Datasets.}
In Table \ref{tab4}, we compare our proposed OH-Former and OH-Former$_{Share}$ with the  state-of-the-art ReID methods. On DukeMTMC-reID, we can see that OH-Former obtains the highest mAP ($82.8\%$) and R-1 ($91.0\%$) with $0.2$–$16.7\%$ and $0.3$–$9.3\%$ improvements compared to other methods, respectively. On Market1501 and MSMT17, OH-Former achieves comparable performance with the latest methods, especially on the R-1 metric. As for the OH-Former$_{Share}$, it surpasses all methods except TransReID by a considerable improvement (at least $+1.9\%$/$+6.6\%$ mAP) on DukeMTMC-reID/MSMT17.

\subsubsection{Results on Occluded Dataset.}
To further explore the generalization ability of ReID models in a complex environment, we utilize the training set of Market1501 to train our model and directly test it on Occluded-Duke dataset. The performance comparison is shown in Table \ref{tab5}. On Occluded-Duke dataset, both OH-Former and OH-Former$_{Share}$ outperform the previous state-of-the-art methods by $1.6$–$23.5$\% and $0.4$–$22.3$\% in mAP, yielding a result of $60.8$\% and $59.6$\% in mAP, respectively. The above fact indicates that our proposed method has outstanding robustness for complex ReID tasks, \eg, occlusion ReID. 

Our model constructs an universal representation that can deal with pedestrians with large variations and even unseen pedestrians as shown in Table \ref{tab5}. Instead of mapping images to a feature space with small intra-identity distance and large inter-identity distance  through learning specific inductive bias, \eg  part correspondence \cite{2017Beyond}, part relations \cite{park2020relation, he2021transreid}, our model embeds person images by considering all-order 
features for both local and non-local relations and learning inductive bias by itself. Thus our omni-relational high-order features are robust to all kinds of variations.


\begin{table}[htbp]
	\centering
	\setlength{\tabcolsep}{16pt}
	\begin{tabular}{ccc}
		\toprule[1pt]
		Method	&	mAP	&R-1	  \\
		\midrule
		PGFA \cite{miao2019pose}	&37.3	&51.4\\
		HOReID \cite{2020High}  &43.8   &55.1\\
		ISP \cite{2020Identity}  &52.3   &62.8\\
		TransReID \cite{he2021transreid} &59.2   &66.4\\
		OH-Former$_{Share}$ (Ours) &59.6 &66.7	\\
		OH-Former (Ours) &\textbf{60.8} &\textbf{67.1} \\
		\bottomrule[1pt]
	\end{tabular}
	\caption{Performance comparison with state-of-the-art methods on Occluded-Duke dataset.}
	\label{tab5}
\end{table}


\section{Conclusion}
In this paper, We present a transformer-based model (OH-Former) for person ReID which takes into account the omni-relational high-order statistics on body tokens, thus making global class and local part tokens discriminative. Furthermore, we propose a Prior Mixing Attention Sharing Mechanism for our OH-Former layer to reduce the computation cost. The local relation perception module introduces inductive biases to augment local information. The effectiveness of each proposed component is sufficiently validated by the ablation studies. The experimental results on various benchmark datasets show that the proposed OH-Former achieves state-of-the-art performance.

\bibliography{aaai22}
\end{document}